\documentclass{article}

    \PassOptionsToPackage{numbers, compress}{natbib}

     \usepackage[preprint]{neurips_2023}

\usepackage[utf8]{inputenc} 
\usepackage[T1]{fontenc}    
\usepackage{hyperref}       
\usepackage{url}            
\usepackage{booktabs}       
\usepackage{amsfonts}       
\usepackage{nicefrac}       
\usepackage{microtype}      
\usepackage{xcolor}         

\usepackage{amsmath}
\usepackage[ruled]{algorithm2e}
\usepackage[capitalize,noabbrev]{cleveref}

\usepackage{standalone}
\usepackage{graphicx}
\usepackage{svg}
\usepackage{rotating}
\usepackage{diagbox}
\usepackage{bbold}
\usepackage{tikz}
\usetikzlibrary{positioning}
\usetikzlibrary{decorations.pathreplacing}
\usetikzlibrary{shapes.geometric}
\usetikzlibrary{calc}
\pgfdeclarelayer{bg}    
\pgfsetlayers{bg,main}

\title{FACT: Federated Adversarial Cross Training}

\author{%
  Stefan Schrod \\
  Department of Medical Bioinformatics\\
  University Medical Center G\"{o}ttingen\\
  G\"{o}ttingen, Germany\\
  \texttt{sschrod@gwdg.de} \\
  \And
  Jonas Lippl \\
  Institute of Theoretical Physics\\
  University of Regensburg\\
  Regensburg, Germany\\
  \texttt{jonas.lippl@student.ur.de} \\
  \And
  Andreas Sch\"{a}fer \\
  Institute of Theoretical Physics\\
  University of Regensburg\\
  Regensburg, Germany\\
  \texttt{andreas.schaefer@physik.uni-r.de} \\
  \And
  Michael Altenbuchinger \\
  Department of Medical Bioinformatics\\
  University Medical Center G\"{o}ttingen\\
  G\"{o}ttingen, Germany\\
  \texttt{maltenb@gwdg.de} \\
}

\begin{document}

\maketitle

\begin{abstract}
Federated Learning (FL) facilitates distributed model development to aggregate multiple confidential data sources. The information transfer among clients can be compromised by distributional differences, i.e., by non-i.i.d. data. A particularly challenging scenario is the federated model adaptation to a target client without access to annotated data. We propose Federated Adversarial Cross Training (FACT), which uses the implicit domain differences between source clients to identify domain shifts in the target domain. In each round of FL, FACT cross initializes a pair of source clients to generate domain specialized representations which are then used as a direct adversary to learn a domain invariant data representation. We empirically show that FACT outperforms state-of-the-art federated, non-federated and source-free domain adaptation models on three popular multi-source-single-target benchmarks, and state-of-the-art Unsupervised Domain Adaptation (UDA) models on single-source-single-target experiments. We further study FACT's behavior with respect to communication restrictions and the number of participating clients.
\end{abstract}

\section{Introduction}
The development of state-of-the-art deep learning models is often limited by the amount of available training data. For instance, applications in precision medicine require a large body of annotated data. Those exist, but are typically distributed across multiple locations and privacy issues prohibit their direct exchange.
Federated Learning (FL) \cite{konevcny2015federated,konevcny2016federated,mcmahan2017communication,smith2017federated} overcomes this issue by distributing the training process rather than sharing confidential data. Models are trained locally at multiple clients and subsequently aggregated at a global server to collaboratively learn. Many FL approaches such as Federated Averaging (FedAvg) \cite{mcmahan2017communication} assume the data to be generated i.i.d., which usually cannot be guaranteed. For instance, different hospitals might use different experimental equipment, protocols and might treat very different patient populations, which can lead to distributional differences between the individual data sites. In such a case, covariate shifts have to be addressed, which is particularly challenging in scenarios where the target client does not have access to labeled training data. There, knowledge has to be both extracted from diverse labeled source clients and transferred without direct estimates of the generalization error.

Unsupervised Domain Adaptation (UDA) \cite{Ganin2015,Tzeng2017,Saito2018} addresses  distributional differences between a label-rich source domain and an unlabeled target domain for improved out-of-distribution performance. Most popular deep learning models utilize an adversarial strategy \cite{Ganin2015, Tzeng2017, Saito2018}, where an adversary is trained to discriminate whether the samples are generated by the source or target distribution. Simultaneously, the feature generator attempts to fool the adversary by aligning the latent representations of the two data sources, which encourages well supported target predictions if the adversary fails to separate the domains. However, these adversarial strategies generally require concurrent access to both source and target data, prohibiting their use in a federated setting without sharing encrypted data representations \cite{peng2019federated} or using artificially generated data \cite{yao2022federated}.

We propose Federated Adversarial Cross Training (FACT), a federated deep learning approach designed to leverage inter-domain differences between multiple source clients and an unlabeled target domain. Specifically, we address the multi-source-single-target setting with non-i.i.d. data sources distributed across multiple clients. To adapt to the domain of an unlabeled target client, we propose to directly evaluate the inter-domain differences between source domains. This allows us to identify domain specific artifacts without adversarial maximization and thus facilitates distributed training among clients. 

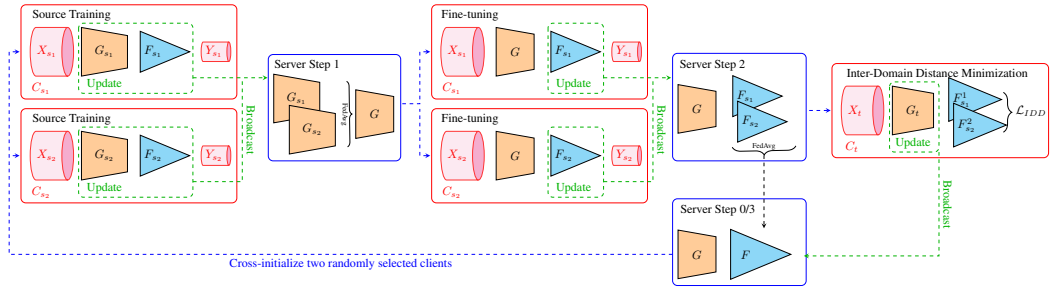
\begin{figure*}[t]
	\begin{center}
		\resizebox{\linewidth}{!}{
				
	\tikzset{%
	    data silo/.style={
    	    cylinder, 
            draw = red!80,
            text = red,
            minimum width = 4em,
            text width=1.7em,
            cylinder uses custom fill, 
            cylinder body fill = magenta!10, 
            cylinder end fill = magenta!40,
            aspect=0.6
        },
        data silox/.style={
    	    cylinder, 
            draw = red!80,
            text = red,
            minimum height = 0.5em,
            cylinder uses custom fill, 
            cylinder body fill = magenta!10, 
            cylinder end fill = magenta!40,
            aspect=0.2
        },
        feature extractor/.style={trapezium,
            draw = black,
            text = black,
            fill = orange!40,
            minimum width = 4em,
            trapezium angle = 80,
            shape border rotate = 270
        },
        head/.style={isosceles triangle,
	        draw,
	        fill=cyan!40,
	        minimum size = 3em
	    },
	    dots/.style={
			draw=none, 
			scale=2.5,
			text height=0.2cm,
			execute at begin node=\color{black}$\vdots$,
		}
	}
	
	\begin{tikzpicture}[x=1.5cm, y=1.5cm, >=stealth]
	
	\node [data silo] (XS1) at (0,4) {$X_{s_1}$};
	\node [feature extractor, right of = XS1, node distance=4.7em] (GS1) {$G_{s_1}$};
	\node [head, right of = GS1, node distance=4em] (HS1) {$F_{s_1}$};
	\node [below of = XS1, xshift=-0.5em, node distance=3em,red] (textSn) {$C_{s_1}$};
	\node [data silox,right of = HS1, node distance=5em] (YS1) {$Y_{s_1}$};
	\node [above of = XS1, xshift=2em, node distance=3em] (textS1) {Source Training};
    
    \begin{pgfonlayer}{bg}    
	\draw[red,rounded corners] ($(XS1.north west)+(-0.2,0.6)$)  rectangle ($(HS1.south east)+(1.4,-0.7)$);
	\end{pgfonlayer}
	
	\draw[black!30!green,thick,dashed,rounded corners] ($(GS1.west)+(-0.05,0.55)$)  rectangle ($(HS1.east)+(0.05,-0.8)$);
	\node [below of = GS1, node distance=2.7em, black!30!green] (YS1) {Update};
	
	\node [data silo] (XS2) at (0,2) {$X_{s_2}$};
	\node [feature extractor, right of = XS2, node distance=4.7em] (GS2) {$G_{s_2}$};
	\node [head, right of = GS2, node distance=4em] (HSj) {$F_{s_2}$};
	\node [below of = XS2, xshift=-0.5em, node distance=3em,red] (textSn) {$C_{s_2}$};
	\node [data silox,right of = HSj, node distance=5em] (YS1) {$Y_{s_2}$};
    \node [above of = XS2, xshift=2em, node distance=3em] (textS2) {Source Training};
    
    \begin{pgfonlayer}{bg}    
	\draw[red,rounded corners] ($(XS2.north west)+(-0.2,0.6)$)  rectangle ($(HSj.south east)+(1.4,-0.7)$);
	\end{pgfonlayer}
	
	\draw[black!30!green,thick,dashed,rounded corners] ($(GS2.west)+(-0.05,0.55)$)  rectangle ($(HSj.east)+(0.05,-0.8)$);
	\node [below of = GS2, node distance=2.7em, black!30!green] (YS1) {Update};

	\def\xsii{4.8};
    \def\ysii{3.1};
    \node [feature extractor, draw = black] (Gserver1_1) at (\xsii,\ysii) {$G_{s_1}$};
	\node [feature extractor, draw = black] (Gserver1_2) at (\xsii+0.3,\ysii-0.6){$G_{s_2}$};
	
	\draw [decorate, decoration = {brace,amplitude=0.4em}] ($(Gserver1_1.north west)+(6em,-0.5em)$) -- node[inner sep=1mm,midway,xshift=-0.2em, rotate=-90] {\tiny FedAvg}  ($(Gserver1_2.south west)+(4.8em,0.5em)$);

	\node [feature extractor, draw = black] (Gserver2) at (\xsii+1.5,\ysii-0.3){$G$};

    \node [] (textServer) at (\xsii+0.2,\ysii+0.65) {Server Step 1};
    \begin{pgfonlayer}{bg}
    \draw [blue,rounded corners] ($(Gserver1_1.north west)+(-0.1,0.4)$)  rectangle (\xsii+2,\ysii-1.2);
    \end{pgfonlayer}

	\node [data silo] (XS1_ft) at (7.9,4) {$X_{s_1}$};
	\node [feature extractor, right of = XS1_ft, node distance=4.7em] (GS1) {$G$};
	\node [head, right of = GS1, node distance=4em] (HS1_ft) {$F_{s_1}$};
	\node [below of = XS1_ft, xshift=-0.5em, node distance=3em,red] (textSn) {$C_{s_1}$};
	\node [data silox,right of = HS1_ft, node distance=5em] (YS1) {$Y_{s_1}$};
	\node [above of = XS1_ft, xshift=1em, node distance=3em] (textS1) {Fine-tuning};

    \begin{pgfonlayer}{bg}    
	\draw[red,rounded corners] ($(XS1_ft.north west)+(-0.2,0.6)$)  rectangle ($(HS1_ft.south east)+(1.4,-0.7)$);
	\end{pgfonlayer}
	
	\draw[black!30!green,thick,dashed,rounded corners] ($(HS1_ft.west)+(-0.05,0.55)$)  rectangle ($(HS1_ft.east)+(0.05,-0.8)$);
	\node [below of = HS1_ft, xshift=0.8em, node distance=2.7em, black!30!green] (YS1) {Update};
	
	\node [data silo] (XS2_ft) at (7.9,2) {$X_{s_2}$};
	\node [feature extractor, right of = XS2_ft, node distance=4.7em] (GS2) {$G$};
	\node [head, right of = GS2, node distance=4em] (HS2_ft) {$F_{s_2}$};
	\node [below of = XS2_ft, xshift=-0.5em, node distance=3em,red] (textSn) {$C_{s_2}$};
	\node [data silox,right of = HS2_ft, node distance=5em] (YS1) {$Y_{s_2}$};
	\node [above of = XS2_ft, xshift=1em, node distance=3em] (textS1) {Fine-tuning};
    
    \begin{pgfonlayer}{bg}    
	\draw[red,rounded corners] ($(XS2_ft.north west)+(-0.2,0.6)$)  rectangle ($(HS2_ft.south east)+(1.4,-0.7)$);
	\end{pgfonlayer}
	
	\draw[black!30!green,thick,dashed,rounded corners] ($(HS2_ft.west)+(-0.05,0.55)$)  rectangle ($(HS2_ft.east)+(0.05,-0.8)$);
	\node [below of = HS2_ft, xshift=0.8em, node distance=2.7em, black!30!green] (YS1) {Update};

	\def\xsiiii{12.5};
    \def\ysiiii{3.5};
    \node [feature extractor, draw = black] (Gserver4) at (\xsiiii,\ysiiii-0.6){$G$};
	\node [head, right of = Gserver4, node distance=4em, yshift =1.05em] (Hserver4_1) {$F_{s_1}$};
    \node [head, right of = Gserver4, node distance=4em, yshift =-1.05em, xshift=0.4em] (Hserver4_n) {$F_{s_2}$};
    
    \node [] (textServer) at (\xsiiii+0.3,\ysiiii+0.25) {Server Step 2};
    \begin{pgfonlayer}{bg}
    \draw [blue,rounded corners] ($(Gserver4.north west)+(-0.1,0.7)$)  rectangle (\xsiiii+2.1,\ysiiii-1.6);
    \end{pgfonlayer}

    \draw [decorate, decoration = {brace,amplitude=0.7em}] ($(Hserver4_n.south west)+(4.8em,-0.9em)$) -- node[inner sep=1mm,midway,yshift=0em] {\tiny FedAvg} ($(Hserver4_1.south east)+(-2em,-3.3em)$);
    \draw [-stealth,thick,dashed] ($(Hserver4_n.south)+(1.1em,-1.5em)$) -- ($(Hserver4_n.south)+(1.1em,-7.5em)$);

    \def\xsiii{12.5};
    \def\ysiii{0.5};
	\node [feature extractor, draw = black] (Gserver3) at (\xsiii,\ysiii-0.4){$G$};
	
	\node [head, right of = Gserver3, minimum width = 4.1em,node distance=4em] (Hserver3) {$F$};
	
	\node [] (textServer3) at (\xsiii+0.4,\ysiii+0.45) {Server Step 0/3};
	\begin{pgfonlayer}{bg}
    \draw[blue,rounded corners] ($(Gserver3.north west)+(-0.1,0.7)$)  rectangle (\xsiii+2.1,\ysiii-1.0);
    \end{pgfonlayer}

	\node [data silo] (XT) at (15.6,2.87) {$X_{t}$};
	\node [feature extractor, right of = XT, node distance=4.5em] (GTn) {$G_{t}$};
	\node [head, right of = GTn, node distance=4em, yshift =0.9em] (HTn0) {$F^1_{s_1}$};
	\node [head, right of = GTn, node distance=4em, yshift =-1.2em, xshift=0.4em] (HTn) {$F^2_{s_2}$};
	\node [below of = XT, xshift=-0.5em, node distance=3em,red] (textT1) {$C_{t}$};
	\node [above of = XT, xshift=6.5em, node distance=3.1em] (textS1) {Inter-Domain Distance Minimization};
	
	\draw [decorate, decoration = {brace,amplitude=0.7em}] ($(HTn0.north west)+(4.5em,-0.5em)$) --  ($(HTn.south west)+(4.1em,0.5em)$);
	\node [right of = XT, node distance=14.2em] (LTn) {$\mathcal{L}_{IDD}$};

    \begin{pgfonlayer}{bg}
	\draw[red,rounded corners] ($(XT.north west)+(-0.2,0.6)$)  rectangle ($(HTn.south east)+(1.35,-0.4)$);
	\end{pgfonlayer}
	\draw[black!30!green,thick,dashed,rounded corners] ($(GTn.west)+(-0.05,0.55)$)  rectangle ($(GTn.east)+(0.1,-0.8)$);
	\node [below of = GTn, node distance=2.7em, black!30!green] (YS1) {Update};
	
	\draw [black!30!green,thick,dashed] ($(HSj.east)+(0.05,-0.5)$) -- ($(HSj.east)+(1,-0.5)$)  -- node[black!30!green,above,rotate=-90] {Broadcast} ($(HSj.east)+(1,1.5)$);
	\draw [-stealth, black!30!green,thick,dashed] ($(HS1.east)+(0.05,-0.5)$) -- ($(HS1)+(2.2,-0.5)$);

	\draw [-stealth, black!30!green,thick,dashed] ($(HS1_ft.east)+(0.05,-0.5)$) -- ($(HS1_ft.east)+(1.35,-0.5)$);
	\draw [black!30!green,thick,dashed] ($(HS2_ft.east)+(0.05,-0.5)$) -- ($(HS2_ft.east)+(1,-0.5)$)  -- node[black!30!green,above,rotate=-90] {Broadcast} ($(HS1_ft.east)+(1,-0.5)$);
	
    \draw[stealth-,thick, blue, dashed] ($(XT.west)+(-0.2,0)$)-- ($(XT.west)+(-0.68,0)$);
    
    \draw[stealth-,thick, blue, dashed] ($(XS1_ft.west)+(-0.2,0)$)-- ($(XS1_ft.west)+(-0.4,0)$)--($(XS1_ft.west)+(-0.4,-1)$)--($(XS1_ft.west)+(-0.75,-1)$);
    \draw[stealth-,thick, blue, dashed] ($(XS2_ft.west)+(-0.2,0)$)-- ($(XS2_ft.west)+(-0.4,0)$)--($(XS1_ft.west)+(-0.4,-1.1)$);
    
    
    \draw [-stealth,thick,black!30!green, dashed] ($(GTn.east)+(0.1,-0.8)$) -- node[black!30!green,above,rotate=-90] {Broadcast} ($(GTn.east)+(0.1,-2.8)$) -- ($(GTn.east)+(-2.5,-2.8)$);
    
    \draw[stealth-,thick,blue, dashed] ($(XS2.west)+(-0.2,0)$) -- ($(XS2.west)+(-0.4,0)$);
    \draw[stealth-,thick,blue, dashed] ($(XS1.west)+(-0.2,0)$) -- ($(XS1.west)+(-0.4,0)$)--($(XS1.west)+(-0.4,-3.9)$)--node[below] {Cross-initialize two randomly selected clients}($(Gserver3.west)+(-0.1,0)$);

	\end{tikzpicture}
	
		}
		\caption{\label{FACT_scheme}Visualization of the FACT training scheme. The server cross-initializes two source clients for \textbf{Source Training}, which locally optimize the model and broadcast it back to the server. The server then aggregates the generator and shares it with the source clients for an additional \textbf{Fine-Tuning} of the domain dependent classification heads. Afterwards, both classification heads and the global generator are sent to the target client for \textbf{Inter-Domain Distance Minimization}. Finally, the optimized generator is broadcasted back to the server and the classification heads are aggregated for the next round of federated training. \label{fig:CFMCD}}
	\end{center}
\end{figure*}

In empirical studies we show that FACT substantially improves target predictions on three popular multi-source-single-target benchmarks with respect to state-of-the-art federated, non-federated and source-free domain adaptation models. Moreover, FACT outperforms state-of-the-art UDA models in several standard settings, also comprising standard single-source-single-target UDA, where the basic model assumptions of FACT are violated (i.e., non-i.i.d. source domains).
Finally, we investigate the behaviour of FACT in different federated learning scenarios to motivate its application to real world problems. We show that FACT benefits from additional source clients even though they are subject to strong covariate shifts, that FACT is stable in applications with many source clients each carrying only a small number of training samples, and that FACT can be efficiently applied in settings with communication restrictions.
Our implementation of FACT, including code to reproduce all results shown in this paper, is publicly available at \url{https://github.com/jonas-lippl/FACT}.

\section{Related Work}
\paragraph{Adversarial Unsupervised Domain Adaptation}
Traditional UDA approaches aim to transfer knowledge from a single labeled source domain to an unlabeled target domain. Proposed methods align the underlying distributions by, e.g., identifying common characteristics of the underlying distributions \cite{ben2010theory,tzeng2014deep,sun2016deep}, by utilizing reconstructed data \cite{ghifary2016deep,zhu2017unpaired} or by using domain adversaries \cite{Ganin2015,Tzeng2017,liu2018unified, Saito2018}. 
With a wide variety of possible applications, the standard single-source-single-target setting has been also extended for additional clients, which is non-trivial for unsupervised target clients \cite{gholami2020unsupervised}. The latter also complicates the weighting between the different domains \cite{ben2010theory}. For a setting with multiple source domains, \cite{zhao2017multiple} proposed an improved DANN model \cite{Ganin2015}, \citet{peng2019moment} proposed moment matching to simultaneously align multiple domains, \citet{ren2022multi} constructed a pseudo target domain between each source-target pair, and \citet{kim2021cds} introduce a self-supervised pre-training scheme to alleviate target performance. This is even further extended by Multi-Source Partial Domain Adaptation (MSPDA) models such as Partial Feature Selection and Alignment (PFSA) \cite{fu2021partial}.
All of these approaches assume centralized data and can not be directly applied to the federated setting.

\paragraph{Federated Learning} FL enables a set of clients to collaboratively learn a prediction model while maintaining all data locally and thus preserving data confidentiality  \cite{mcmahan2017communication,bonawitz2017practical,konevcny2015federated,konevcny2016federated}. This can be achieved by encrypting the data \cite{gilad2016cryptonets,mohassel2017secureml} or by sharing locally trained models to aggregate knowledge. Typically, a global model is constructed by averaging the individual clients' models, e.g., via Federated Averaging (FedAvg) \cite{mcmahan2017communication}. For non-i.i.d. data, model aggregation can remove client specific characteristics of the data, leading to negative transfer and thus suboptimal performance \cite{zhao2018federated,acar2021federated}. Approaches designed for non-i.i.d. data sources adapt distributed multi-task learning \cite{smith2017federated}, use secure transfer learning \cite{liu2020secure} or inter client domain generalization \cite{zhang2021federated,liu2021feddg}. Furthermore, improved aggregation strategies for non-i.i.d data were proposed, which consider the client specific momentum \cite{hsu2019measuring} or dynamically regularize the clients \cite{acar2021federated} to align the optimal solutions of global and local models.

\paragraph{Unsupervised Federated Domain Adaptation}
Unlabeled data naturally occur in many applications of FL. In the special case of completely unlabeled target data, UDA can enhance the out-of-distribution performance provided that the data access restrictions are resolved. Seminal work by \citet{peng2019federated} introduced Federated Adversarial Domain Adaptation (FADA), which uses local feature extractors to generate privacy preserving representation that can be shared for distribution alignment. Similarly, Federated Knowledge Alignment (FedKA) \cite{sun2023feature} attempt to align local feature extractors to a global data embedding provided by the cloud using the Multiple Kernel Maximum Mean Discrepancy. Related approaches were recently applied to MRI data \cite{li2020multi,guo2021multi,zeng2022gradient}, demonstrating the benefits of multi-institutional collaborations.

A second closely related line of work is source-free domain adaptation \cite{li2020model,ahmed2021unsupervised,liang2020we,liu2021source,kurmi2021domain,xia2021adaptive,shenaj2023learning}. Here, a pre-trained source model is used to transfer knowledge to the target domain and subsequently fine-tuned exclusively on the target data with the objective to make the target predictions both individually certain and diverse \cite{liang2020we}. Accordingly, only the pre-trained source model needs to be shared, which guarantees data confidentiality. This approach is basically the one-shot federated setting requiring only a single round of federated communication. In particular, \citet{liang2020we} and \citet{ahmed2021unsupervised} addressed a possible extension to the multi-source-single-target setting by combining the outcome of multiple single-source-single-target models. These models additionally use image augmentation strategies on the source data, which on their own out-perform state-of-the-art domain adaptation settings on multi-source-single target experiments \cite{liang2020we,ahmed2021unsupervised}. One should note that this is in contrast to standard UDA and multi-source UDA approaches that attempt to eliminate domain differences algorithmically without relying on augmented training samples.

\section{Method}
In this work, we address the federated multi-source-single-target setting where the data is both non-i.i.d. and distributed across multiple clients. We assume each client to be associated with an individual domain and focus on the out-of-distribution performance on data from a single target client without training labels, corresponding to the Unsupervised Federated Domain Adaptation (UFDA) setting  introduced by \citet{peng2019federated}. Let $S$ be a global server managing the training process for a set of local source clients $\mathcal{C}_s$ and let $X\in \mathcal{X}$ be the features associated with $K$ classification labels $\{1,2,\ldots,K\}=Y\in \mathcal{Y}$. The data are distributed across $n$ source clients $C_{s_1},\ldots,C_{s_n}\in \mathcal{C}_s$, associated with annotated training data $(X_{s_i},Y_{s_i})$ and a target client $C_{t}$ that holds solely covariate data $(X_{t})$ not associated with any labels. We assume that the data of all clients share the same input and output $\mathcal{X}\times \mathcal{Y}$.

\subsection{Motivation: Exploiting inter-domain differences}
Following the general concept of adversarial learning strategies, we attempt to
learn a domain invariant data representation to increase the support of target samples. This is usually achieved by evaluating whether a sample is generated by the source or the target distribution and by suppressing domain specific features. This requires, however, an additional adversarial maximization step to identify such features. 

FL naturally enables the access to multiple source domains exhibiting individual characteristics. FACT assumes the data of each client to be non-i.i.d. and uses these implicit inter-domain differences to evaluate the support of an independent and unlabeled target. Let $(X_1,Y_1)\sim P_1$ and $(X_2,Y_2)\sim P_2$ be generated by two distributions subject to strong domain shifts. Then, two independently trained models $F_1$ and $F_2$ on $X_1$ and $X_2$, respectively, will be biased towards their corresponding source data. This encourages inconsistent classifications for an independent target domain $X_t\sim P_t$ and thus the difference in classifications $\hat{Y}^1_t$ and $\hat{Y}^2_t$ provides a measure of covariate shifts. Thus, using the two source specific models, we define the adversary via the Inter-Domain Distance (IDD):
 \begin{equation}
     \begin{aligned}
     \min_{G}\,  &\mathcal{L}_{\textrm{IDD}}(X_t) \label{eq:MCDstep2}\\
     \mathcal{L}_{\text{IDD}}(X_t) = E_{x_t\sim X_t} &\lVert{F_1(G(x_t))-F_2(G(x_t))}\rVert_1\,,
     \end{aligned}
 \end{equation}
which minimizes the domain specific artifacts with respect to a latent data representation $G(X)$ and as such it enforces agreement between target estimates. The IDD objective is inspired by the discrepancy loss introduced for the Maximum Classifier Discrepancy (MCD) adversary by \citet{Saito2018}. There, in contrast to our proposed inter-domain training, the classifiers are trained on data generated by a single domain. Hence, an adversarial maximization step is essential to bias the two classifications towards contradicting results on the target data. In this step concurrent access to both source and target data is required to account for an uncontrolled loss of target accuracy, restricting the federated application of MCD.

\begin{algorithm}[t]
    \caption{FACT: Federated Adversarial Cross Training \label{alg:FACT}}
    \SetKwInOut{Input}{Input}
    \Input{The round of federated learning $r$, a Global server model $G^r$ and $F^r$, a set of Source clients $\mathcal{C}_s$ and a target client $C_t$ associated with private data $(X_s,Y_s)$ and $(X_t)$, respectively.}
    \For{$r <$ max rounds}{
    \textbf{Server Step 0:} Cross-initialize two randomly sampled source clients $C^r_{s_1}, C^r_{s_2}\sim\mathcal{C}_s$ with the global $G^r$ and $F^r$.\\
    \textbf{Source Training:} Update the local models $F^r_{s_i}(G^r_{s_i})$ by $\nabla_{G^r_{s_i},F^r_{s_i}}\mathcal{L}_{ce}(X_s,Y_s)$.\\
    \textbf{Server Step 1:} Collect and aggregate the feature extractors $G'=1/2(G^r_{s_1}+G^r_{s_2})$ and broadcast $G'$ back to the source clients.\\
    \textbf{Fine-tuning:} Optimize the classification heads to fit frozen $G'$ by $\nabla_{F^r_{s_i}}\mathcal{L}_{ce}(X_s,Y_s)$.\\
    \textbf{Server Step 2:} Together with $G'$ share the fine-tuned $F_{s_1}$ and $F_{s_2}$ with the target client $C_t$.\\
    \textbf{Inter-Domain Distance Minimization:} Update the feature generator $G'$ to minimize the IDD by $\nabla_{G'}\mathcal{L}_{\textrm{IDD}}(X_t)$.\\
    \textbf{Server Step 3:} Collect the updated $G':=G^{r+1}$ and aggregate $F^{r+1}=1/2(F_{s_1}+F_{s_2})$
    }
\end{algorithm}

\subsection{Federated Adversarial Cross Training}
We propose Federated Adversarial Cross Training (FACT) as a simple and highly efficient training scheme which leverages the inter-domain differences of distributed data to maximize the information transfer to an independent target domain without access to labeled data. The basic concept of FACT is summarized in Figure \ref{FACT_scheme} and outlined in the following.

Model training is managed by a global server $S$, responsible to distribute, collect and aggregate the individual client models. FACT consists of a global feature generator $G$ which is optimized to yield domain invariant representations for all of the domain specific classification heads $F_{s_i}$, for all source domains $s_i$. At the beginning of the training process the global model generator $G$ and classifier $F$ are initialized at the server.
Each round of federated learning $r$ is split into three main steps: source training, source fine-tuning and target inter-domain distance minimization, which are applied iteratively.

\paragraph{Source Training} At the beginning of each round $r$ the server $S$ cross-initializes two randomly selected source clients $C^r_{s_1},C^r_{s_2} \in \mathcal{C}_s$ with the global model $(G^r,F^r)$. Each client $C^r_{s_i}$ updates the full model $(G^r,F^r)$ to fit their respective source data using a standard Cross-Entropy objective
\begin{equation}
\begin{aligned}
    \min_{G_{s_i},F_{s_i}} \mathcal{L}_{ce} (X_{s_i},Y_{s_i})\\
    \mathcal{L}_{ce}(X_s,Y_s)=-E_{(x_s,y_s)\sim(X_s,Y_s)}&\sum^{K}_{k=1}\mathbb{1}_{[k=y_s]}\log p(y|x_s)\,. \label{eq:FACT_source}
    \end{aligned}
\end{equation}
The updated feature extractor is subsequently broadcasted back to the server to update the global feature extractor $G'$:
\begin{equation}
    G' \leftarrow \sum_{C^r_{s_i}}\frac{1}{2} G^r_{s_i}\,.
\end{equation}

\paragraph{Source Fine-Tuning}
The aggregation of the global feature extractor changes also the latent representations of the classification heads. This can potentially compromise model performance. To compensate this effect and to further bias the models towards the individual source data, we fine-tune the classification heads $F_{s_1}, F_{s_2}$ to fit the aggregated feature extractor $G'^{r}$:
\begin{equation}
    \min_{F_{s_i}} \mathcal{L}_{ce} (X_{s_i},Y_{s_i})\,. \label{eq:FACT_fine}
\end{equation}

\paragraph{Inter-Domain Distance Minimization}
The two fine-tuned classification heads together with the feature extractor are then transferred to the target client, where the individual model representations are used to quantify domain differences. Those are mitigated by minimizing the IDD loss
\begin{equation}
    \min_{G_t}  \mathcal{L}_{\textrm{IDD}}(X_t)
\end{equation}
with respect to the joint feature generator. Thus, the feature space is updated such that domain differences are mitigated on the target domain. The updated $G_{r+1}\leftarrow G_t$ is subsequently broadcaster back to the server, where 
a new global classification head is aggregated and the next round of federated training is initiated.

By design, the inter-domain distance provides a direct measure for the support of the target domain. Hence, we select the final model based on minimal IDD in all of the experiments. For practical applications this can speed up the training significantly by reducing the number of trained epochs and client communications.
The full training procedure is shown in \cref{alg:FACT}. Note that the fine-tuning step can increase communication costs substantially, since it requires an additional transfer of the commonly large generator to the sources clients. Thus, in subsequent experiments, we present results for both FACT and a simplified version of FACT which does not perform the fine-tuning step (FACT-NF). 

\begin{table*}[t]
    \centering
    \caption{Target client accuracy obtained for four source clients on Digit-Five data. Best performing methods within error bars are highlighted.}
    \begin{tabular}{lllllll}
        \toprule
        Models & $\rightarrow$MNISTM  & $\rightarrow$MNIST  &$\rightarrow$SVHN    &$\rightarrow$SYN    & $\rightarrow$USPS    & Avg      \\
        \midrule
        MCD                 & $72.5\pm0.7$      & $96.2\pm0.8$      & $78.9\pm0.8$      & $87.5\pm0.7$      & $95.3\pm0.7$      & 86.1 \\
        M$^3$SDA-$\beta$    & $72.8\pm1.1$      & $98.4\pm0.7$      & $81.3\pm0.9$      & $89.6\pm0.6$      & $96.1\pm0.8$      & 87.7 \\
        MDDA                & $78.6\pm0.6$      & $98.8\pm0.4$      & $79.3\pm0.8$      & $89.7\pm0.7$      & $93.9\pm0.5$      & 88.1 \\
        LtC-MSDA &$85.6\pm0.8$& $\mathbf{99.0\pm0.4}$ & $83.2\pm0.6$& $93.0\pm0.5$& $\mathbf{98.3\pm0.4}$ & 91.8\\
        PFSA& $89.6\pm1.2$& $\mathbf{99.4\pm0.1}$ & $84.1\pm1.1$& $\mathbf{95.7\pm0.3}$ & $\mathbf{98.6\pm0.1}$ & 93.5 \\
        \midrule
        FADA               & $62.5\pm0.7$      & $91.4\pm0.7$      & $50.5\pm0.3$      & $71.8\pm0.5$      & $91.7\pm1.0$      & 73.6 \\
        FedKA                  & $77.3\pm1.0$      & $96.4\pm0.2$      & $13.8\pm0.8$      & $79.5\pm0.7$      & $96.6\pm0.4$      & 72.7 \\
        \midrule
        FACT-NF  & $89.6 \pm 0.7$ & $\mathbf{99.2 \pm 0.1}$ & $88.8 \pm 0.3$ & $95.0 \pm 0.1$ & $\mathbf{98.5 \pm 0.2}$ & 94.2 \\ 
        FACT                               & $\mathbf{92.9\pm0.7}$      & $\mathbf{99.2\pm0.1}$      & $\mathbf{90.6\pm0.4}$      & $95.1\pm0.1$      & $\mathbf{98.4\pm0.1}$      & $\mathbf{95.2}$ \\
        \bottomrule
    \end{tabular}
    \label{tab:digitfive_single_target}
\end{table*}

\section{Experiments}
We evaluate a variety of different federated scenarios with respect to the fraction of correctly predicted classifications for an independent target domain. First, we show the multi-source-single-target setting where each client is associated with an individual domain and a large dataset. Second, we evaluate FACT for the standard single-source-single-target UDA setting by splitting a single domain across two clients, which violates FACT's assumption of existing domain differences. Finally, we analyze the behavior of FACT in different federated learning scenarios, studying (1) the negative transfer in case of suboptimal source data, (2) the number of participating clients while limiting the data present at each individual location, and (3) the effect of reducing the number of communication rounds.

\paragraph{Datasets} We use three popular benchmark datasets for multi-source-single-target adaptation, namely Digit-Five \cite{peng2019moment}, Office \cite{saenko2010adapting} and Office-Caltech10 \cite{gong2012geodesic}. Each of them is known to exhibit strong domain shifts: Digit-Five combines five independently generated digit recognition datasets (MNIST, MNIST-M, USPS, Street-View House Numbers (SVHN), synthetic digits (SYN)). Office consists of 31 classes of office appliances collected form three different sources (Amazon.com, Webcam, DSLR pictures). Office-Caltech10 \cite{gong2012geodesic} adds Caltech-265 as an additional domain to the Office dataset, resulting in a total of 10 shared classes between all domains. In line with previous work \cite{peng2019moment,peng2019federated,liang2020we,ahmed2021unsupervised}, we use an independent target test set for Digit-Five and use the unlabeled data for evaluation for Office and Office-Caltech10. The number of the used train and test samples are listed in the Supplementary Materials.

\paragraph{Baselines}
We compare FACT to multiple types of state-of-the-art baselines, comprising both federated and non-federated domain adaptation models. The federated baselines are FADA \cite{peng2019federated}, which calculates a DANN based adversary for each domain pair at the server based on an encrypted representation of the private data, and Federated Knowledge Alignment (FedKA) \cite{sun2023feature}, which matches the feature embedding between two domains using a multiple kernel variant of the Maximum Mean Discrepancy in combination with a federated voting strategy for model fine-tuning. Even though data access restrictions prohibit the application of MCD \cite{Saito2018} and alike UDA models, we include the less restrictive non-federated benchmarks where all data are available at the same location and can be concurrently accessed. Here, we compare to a multi-source adaptation of MCD \cite{Saito2018} and to M$^3$SDA-$\beta$ as implemented by \citet{peng2019moment}, Multi-source Distilling Domain Adaptation (MDDA) \cite{zhao2020multi}, Learning to Combine for Multi-Source Domain Adaptation (LtC-MSDA) \cite{wang2020learning} and Partial Feature Selection and Alignment (PFSA) \cite{fu2021partial}. The performances of all competing methods were extracted from the respective original research article, with the exception of the MCD results which were obtained from \cite{peng2019federated} and \cite{fu2021partial}.

\paragraph{Implementation details} Our implementation of FACT is based on PyTorch \cite{paszke2019pytorch} and is publicly available at \url{https://github.com/jonas-lippl/FACT}. The federated setting was simulated on a single machine to speed up computation for all experiments. All calculations were performed on a cluster with 8 NVIDIA A100 gpus and 256 cpu cores. Digit-Five was trained from scratch using randomly initialized layers and for Office and Office-Caltech10, we initialized the feature generator using a pre-trained ResNet101 \cite{he2016deep}.
Overall, the chosen architecture follows the design choices of \citet{peng2019federated} to guarantee a fair comparison. The results were averaged over 10 repeated runs for Digit-Five and 5 repeated runs for Office and Office-Caltech10. We used a batch-size of 128 and a learning rate scheduler $\eta = \eta_0 \cdot (1+10\cdot p)^{-0.75}$ as proposed by \citet{ganin2016domain} with an initial learning rate of $\eta_0=0.005$ for the multi-source-multi-target experiments and $\eta_0=0.01$ for the single-source-single-target experiments to guarantee convergence. For additional details we refer to the Supplementary Materials.

\begin{table}[t]
    \centering
    \caption{Target client accuracy obtained for using two source clients on the Office dataset. (A: Amazon, D:DSLR, W:Webcam)}
    \begin{tabular}{lllll}
        \toprule
        Method   & $\rightarrow$ A   &$\rightarrow$ D & $\rightarrow$ W  &Avg\\
        \midrule
        MCD & 54.4 & 99.5 & 96.2 & 83.4\\ 
        M$^3$SDA & 55.4 & 99.4 & 96.2 & 83.7\\
        MDDA & 56.2 & 99.2 & 97.1 & 84.2\\
        LtC-MSDA & 56.9 & 99.6 & 97.2 & 84.6\\
        PFSA & 57.0 &$\mathbf{99.7}$ & $\mathbf{97.4}$ & 84.7\\
        \midrule
        FACT-NF & 69.1  & 99.3  & 95.5  & 88.0 \\ 
        FACT & $\mathbf{70.5}$  & 99.5  & 96.0  & $\mathbf{88.7}$ \\
        \bottomrule
    \end{tabular}
    \label{tab:office}
\end{table}

\subsection{Multi-source-single-target with distinct source domains}
As a baseline, we evaluate the standard federated multi-source-single-target setting \cite{peng2019moment,peng2019federated,sun2023feature,liang2020we,ahmed2021unsupervised}, where a small number of clients each holds a distinct domain with a large number of available training samples. The results obtained for Digit-Five are shown in \cref{tab:digitfive_single_target}. We observe that FACT outperforms the competing models with an average accuracy of 95.2\%, which is followed by PFSA with 94.0\%. This performance gain is mainly attributed to the scenario with SVHN and MNISTM as target domain. On the former, FACT achieved an accuracy of 90.6\%, which is substantially better than for all competing methods with accuracies ranging from 13.8\% (FedKa) to 84.1\% (PFSA). Considering the remaining target client scenarios, FACT shows comparable performance to LtC-MSDA and PFSA, and substantially better performance than the other methods, in particular compared to federated baselines FADA and FedKA.

For the Office data (\cref{tab:office}) no federated baselines were available, hence, we only compare to less restrictive non-federated models. We observe that FACT significantly outperforms the average performance of all non-federated baselines. This is mainly attributed to the increased accuracy on the Amazon target to 70.5\%, which might be attributed to the efficient aggregation of the very similar DSLR and Webcam domains. FACT shows slightly compromised performance on the Webcam domain with 96.0\% accuracy compared to the best performing model PFSA with 97.4\%.

Considering Office-Caltech10 (\cref{tab:office_caltech_single_target}), FACT outperforms both the federated and non-federated baselines for three out of four possible targets and at average. 

For all three experiments, we observe that the fine-tuning step improves model performance, as becomes evident from comparing FACT with FACT-NF. This particularly holds true for the more challenging target domains, such as Amazon for the Office dataset or MNISTM and SVHN for the Digit-Five dataset. For the remaining scenarios, FACT-NF closely matches the results of FACT.

\begin{table}[t]
    \centering
    \caption{Target client accuracy obtained for three source clients on the Office-Caltech10 dataset (C: Caltech, A: Amazon, D:DSLR, W:Webcam).}
    \begin{tabular}{llllll}
        \toprule
        Method                  &$\rightarrow$ C  & $\rightarrow$ A   &$\rightarrow$ D & $\rightarrow$ W  &Avg\\
        \midrule
        MCD                    &91.5           &92.1           &99.1           &$\mathbf{99.5}$           &95.6\\
        M$^3$SDA-$\beta$        &92.2           &94.5           &99.2           &$\mathbf{99.5}$           &96.4\\
        \midrule
        FADA                     &88.7   &84.2   &87.1   &88.1&87.1\\
        FedKA                    &39.7  &59.9  &30.2  &23.4  &38.3\\
        \midrule
        FACT-NF        & 95.0  & $96.1$ & 99.1  & 98.3  & 97.1 \\ 
        FACT               & $\mathbf{95.5}$ & $\mathbf{96.3}$ & $\mathbf{99.4}$  & 99.0  &$\mathbf{97.6}$ \\
        \bottomrule
    \end{tabular}
     \label{tab:office_caltech_single_target}
\end{table}

\subsection{FACT improves on baselines even if model assumptions are violated}
FACT uses inter-domain differences between source domains to identify and mitigate domain shifts in the target domain. Next, we evaluate the performance of FACT in a scenario where the data are approximately i.i.d. and as such violate the assumptions. For this, we consider the standard UDA single-source-single-target scenario. Since FACT requires a minimum of two source clients to apply the proposed training scheme without limitations, we artificially distribute each source domain to two equally sized source clients. \cref{tab:single_source_uda} shows the obtained target accuracies for three popular test settings on Digit-Five compared to state-of-the-art results obtained by Conditional Adversarial Domain Adaptation with Entropy Conditioning (CDAN+E) \cite{long2018conditional}, MCD \cite{Saito2018}, Cluster Alignment with a Teacher (CAT) \citet{deng2019cluster}, and robust RevGrad with CAT (rRevGrad+CAT) \citet{deng2019cluster}.

We observe that FACT outperforms the state-of-the-art baselines in two out of three commonly tested scenarios for Digit-Five. For SVHN to MNIST, FACT shows a compromised accuracy of 90.6\% compared to 98.8\% for the best performing model (rRevGrad+CAT). However, for USPS to MNIST and vice versa, FACT obtained the highest accuracy of all baselines. These finding support FACT's capability to establish domain robust models even in scenarios where no or little domain differences are present among the source domains.

\begin{table}[t]
    \centering
    \caption{Target accuracies obtained for the standard single-source-single-target UDA setting (S: SVHN, M:MNIST, U: USPS).}
    \begin{tabular}{lllll}
        \toprule
         & S$\rightarrow$M & U$\rightarrow$M& M$\rightarrow$U & Avg\\
        \midrule
        CDAN+E  & 89.2 & 98.0 & 95.6 & 94.3\\
        MCD  & 96.2 & 94.1 & 94.2 & 94.8\\
        MCD+CAT& 97.1& 96.3& 95.2& 96.2\\
        rRevGrad+CAT & \textbf{98.8} & 94.0 & 96.0 & \textbf{96.3}\\
        \midrule
        FACT & 90.6  & \textbf{98.6}  & \textbf{98.8}  & 96.0\\
        \bottomrule
    \end{tabular}
    \label{tab:single_source_uda}
\end{table}

\subsection{FACT is robust with respect to suboptimal source domains}
The results from the previous section show that the accuracy obtained for different domain combinations can differ significantly, dependent on the domain shift between the source and the target or the sample sizes. Thus, the combination of heterogeneous source domains might not be beneficial in general but could also introduce negative transfer \cite{wang2019characterizing} and could compromise the multi-source predictions. Negative transfer is a common issue for standard model aggregation with FedAvg, since opposing updates for highly homogeneous data sources might cancel each other, leading to slowly or non-converging models \cite{mcmahan2017communication,li2020federated}.

Starting from the single-source setting of the previous section we gradually increase the number of participating source domains and evaluate the target accuracy for all possible combinations thereof. The results for Digit-Five are shown in \cref{fig:combined_plot}a. We observe that each added source domain improves the average accuracy for each target. In particular, SVHN, MNISTM and SYN substantially benefit from additional source clients, which might be due to the large domain shifts these domains exhibit to most of the potential source domains. However, we observe that specific source-target combinations are already able to closely match the performance of FACT using all of the four available sources, i.e., SYN $\rightarrow$ SVHN with 92.7\%, SYN $\rightarrow$ MNISTM with 88.3\% and SVHN $\rightarrow$ SYN with 95.7\% accuracy, respectively (for the full results of all source-target combinations see the Supplementary Materials). In all three cases, adding additional source clients did not significantly compromise the overall predictions and the small errors of the full model suggest stable model performance. Thus, many source clients might be beneficial in general and minimizing IDD might prevent negative transfer by actively encouraging invariant domains.

\subsection{FACT generalizes to many clients}
In contrast to the previous experiments, we demonstrate that FACT can efficiently deal with an increased number of source clients. We use all source domains and evenly split each source across 3, 5 and 10 individual clients for a total of 12, 20 and 40 source clients, respectively. Note that the total number of training samples remains constant and only the sample size at each location is reduced and as such it becomes more likely that data from the same domain are used in cross training. We observe (\cref{fig:combined_plot}b) that the number of clients only marginally impacts the target performance, especially for the USPS and MNIST, suggesting the applicability of FACT also in scenarios with many sources which individually contribute only small-sized data. Noteworthy, the obtained results consistently outcompete the federated and non-federated baselines evaluated without splitting the data into multiple sources.

\subsection{FACT's performance under communication restrictions}
\begin{figure*}
    \centering
    \includegraphics[width=\textwidth]{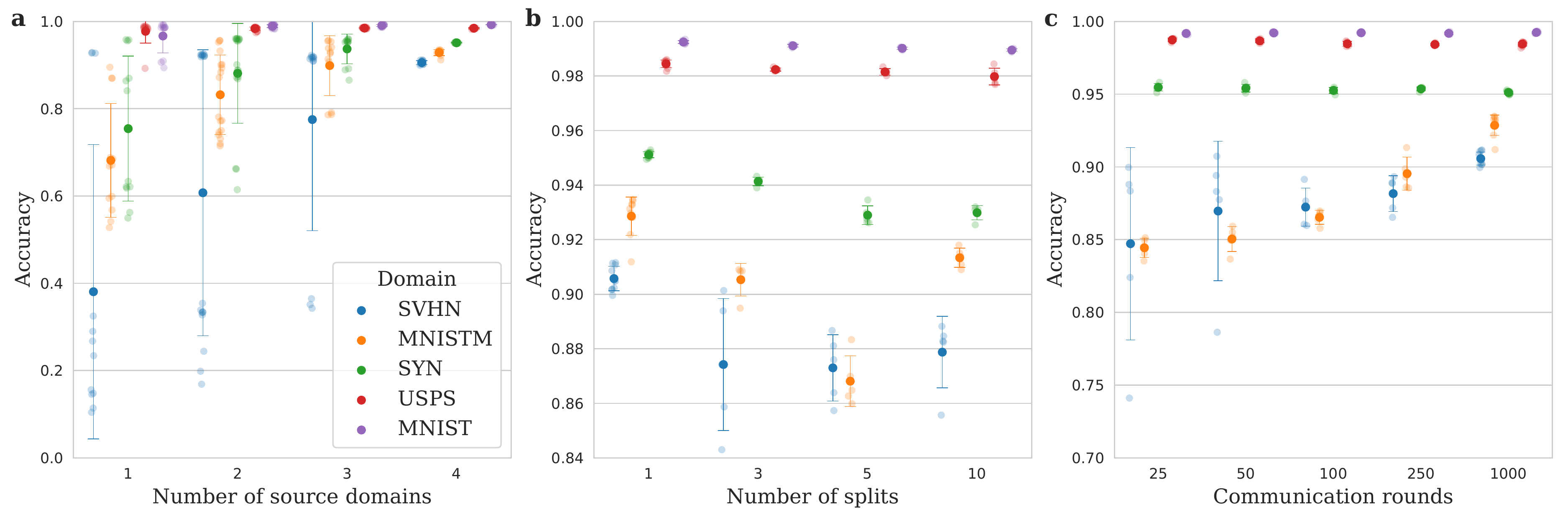}
    \caption{Target client accuracies obtained on Digit-Five data for \textbf{(a)} a varying number of participating source domains evaluated for all possible domain combinations, \textbf{(b)} additional source clients where the individual source domains are split across multiple clients, and \textbf{(c)} communication restrictions where the number of communication rounds are varied for a fixed number of  training epochs.}
    \label{fig:combined_plot}
\end{figure*}

Frequent communication with a server can render federated learning approaches impractical. Thus, we systematically study FACT's performance with respect to the number of communication rounds. For this, we  fix the number of totally trained epochs to 1000 but vary the number of communication rounds.
On Digit-Five, we observe (\cref{fig:combined_plot}c) no significant change in accuracy for the three best performing target domains, i.e., MNIST, SYN and USPS, even for as little as 25 rounds of FL. The performance of the more challenging target domains, MNISTM and SVHN, were  gradually decreased to 84.4\% and 84.7\%, respectively. There, however, estimated accuracies showed a high variance, suggesting that frequent communication might be beneficial but is not necessary in general. Moreover, the obtained mean accuracy of all targets is still substantially better than those by the federated and non-federated baselines even for highly restricted communication.

\section{Conclusion}
We showed that inter-domain distances can be directly leveraged for multi-source-single-target domain adaptation, which also holds true for federated scenarios where privacy issues prohibit direct data exchange between clients. FACT combines only two objectives, namely minimizing a standard classification loss with minimizing inter-domain distances. Although simple, this approach substantially outperforms the state-of-the-art in the majority of the performed experiments.

This also applies to standard UDA scenarios with a single source domain, violating the main assumption of FACT that data available at different clients are non-i.i.d.. We further showed that a suboptimal selection of source clients has only a minor effect on target accuracy. Further, FACT performs well in scenarios where data are distributed across multiple source clients and where server communication is restricted. In summary, FACT is a simple, highly efficient and highly competitive approach which might be valuable to many federated real world scenarios.

\paragraph{Limitations}
We could empirically show that controlling the inter-domain distance among both non-i.i.d. and i.i.d. source clients substantially improves multi-source-single-target domain adaptation. Yet, however, there are no theoretical guarantees to which scenario this concepts applies and to what extend.
In particular, for the i.i.d. setting where only sample domain differences might be observed, it may be hard to verify that the identified domain differences are significant enough to guarantee target adaptation.
Since the distributional alignment is based on the joint feature extractor, each client requires sufficient resources to calculate the respective gradients, potentially restricting the applicability of FACT on IoT devices for large models. Similarly, transferring large models between the clients may increase the training time. In this case, reducing the number of communication rounds or neglecting the fine-tuning could be an option. 

\section*{Acknowledgements}
This work was supported by the German Federal Ministry of Education and Research (BMBF) [01KD2209D].

\bibliographystyle{plainnat}
\bibliography{FL}

\newpage
\section*{Supplementary Materials}
\subsection*{Implementation Details}
For Digit-Five \cite{peng2019moment}, we use a feature generator consisting of two convolutional layers and a classifier consisting of three fully connected (FC) layers. The detailed architecture can be found in \cref{tab:model_architecture}. For the Office \cite{saenko2010adapting} and Office-Caltech10 \cite{gong2012geodesic} data, the generator is provided by a pre-trained ResNet101 \cite{he2016deep} and the classifier consists of two FC layers. We additionally use Batch-Normalization (BN) layers and ReLu activation functions as outlined in \cref{tab:model_architecture}. For all experiments, the maximum number of epochs trained at the clients is set to 1000 with a batch-size of $B=128$. Further, we use an initial learning rate of $\eta_0 = 0.005$ for all experiments, except for the single-source-single-target setting. In the latter, we use a higher initial  learning rate of $\eta_0 = 0.01$ to guarantee convergence within the 1000 training epochs.
\begin{table}[h]
    \centering
    \caption{Detailed model architecture used for Digit-Five and the classifier used in combination with the pre-trained ResNet101 for Office and Office-Caltech10.}
    \label{tab:model_architecture}
    \begin{tabular}{ll}
    \toprule
    layer & configuration \\
    \midrule
    \multicolumn{2}{c}{Feature Generator (Digit-Five)}\\
    \midrule
    1 & Conv2D (3, 64, 5, 1, 2), BN, ReLU, MaxPool\\
    2 & Conv2D (64, 128, 5, 1, 2), BN, ReLU, MaxPool\\
    \midrule
    \multicolumn{2}{c}{Classifier (Digit-Five)} \\
    \midrule
    1 & DropOut (0.5), FC (8192, 3072), BN, ReLu\\
    2 & DropOut (0.5), FC (3072, 100), BN, ReLu\\
    3 & DropOut (0.5), FC (100, 10), BN, Softmax\\
    \midrule
    \multicolumn{2}{c}{Classifier for ResNet101}\\
    \midrule
    1 & DropOut (0.5), FC (1000, 500), BN, ReLU \\
    2 & FC (500, number of classes), BN, ReLu, Softmax\\
    \bottomrule
    \end{tabular}
\end{table}

\begin{table}
    \centering
    \caption{Number of samples for each dataset and domain.}
    \label{tab:number_of_samples}
    \begin{tabular}{cc c c c c c c}
    \multicolumn{8}{c}{Digit-Five}\\
    \midrule
    Split & MNISTM & MNIST & SVHN & SYN & USPS & & Total\\
    \midrule
    Train & 25000 & 25000 & 25000 & 25000 & 7348 & & 107348\\
    Test & 9000 & 9000 & 9000 & 9000 & 1860 & & 37860\\
    \midrule
    \multicolumn{8}{c}{Office}\\
    \midrule
    Split & & Amazon &  & DSLR & Webcam & & Total\\
    \midrule
    Total & & 2817 &  & 498 & 795 & &  4110\\
    \midrule
    \multicolumn{8}{c}{Office-Caltech10}\\
    \midrule
    Split & & Amazon & Caltech & DSLR & Webcam & & Total\\
    \midrule
    Total & & 958 & 1123 & 157 & 295 & &  2533\\
    \bottomrule
    \end{tabular}
\end{table}

\begin{table}
    \centering
    \caption{Number of classes for each dataset}
    \label{tab:number_of_classes}
    \begin{tabular}{cc}
    Dataset & Number of classes \\
    \midrule
    Digit-Five & 10 \\
    Office & 31 \\
    Office-Caltech10 & 10 \\
    \end{tabular}
\end{table}

\subsection*{Single-source-single-target UDA}
\cref{tab:two_source_same_domain} shows the target accuracies obtained by FACT for all possible source-target combinations on the Digit-Five data. The results are averaged over 5 repeated runs. Further note that in order to run FACT without modifications, we equally split the source data between two source clients. As such, the distributional differences between the two sources are sample dependent, which potentially decreases target convergence. Therefore, we used a higher initial learning rate $\eta_0 = 0.01$ for these experiments.

\begin{table}
    \centering
    \caption{Target accuracies obtained for the standard single-source-single-target UDA setting for all possible source-target combinations of the Digit-Five data.}
    \begin{tabular}{llllll}
        \diagbox{source}{target}   & $\rightarrow$ MNISTM & $\rightarrow$ MNIST & $\rightarrow$ SVHN & $\rightarrow$ SYN & $\rightarrow$ USPS\\ 
        \midrule
        MNISTM & -  & 99.17 $\pm$ 0.04 & 29.0 $\pm$ 2.52 & 85.86 $\pm$ 0.96 & 98.75 $\pm$ 0.11\\ 
        MNIST & 61.27 $\pm$ 2.85 & -  & 16.91 $\pm$ 3.28 & 64.41 $\pm$ 6.05 & 98.75 $\pm$ 0.11\\ 
        SVHN & 54.62 $\pm$ 1.31 & 90.58 $\pm$ 0.61 & -  & 95.74 $\pm$ 0.13 & 92.56 $\pm$ 4.51\\ 
        SYN & 88.28 $\pm$ 1.07 & 98.54 $\pm$ 0.07 & 92.65 $\pm$ 0.25 & -  & 98.2 $\pm$ 0.25\\ 
        USPS & 67.46 $\pm$ 1.02 & 98.58 $\pm$ 0.04 & 12.19 $\pm$ 1.97 & 60.85 $\pm$ 2.31 & - \\ 
    \end{tabular}
    \label{tab:two_source_same_domain}
\end{table}

\end{document}